%% file: main.tex
\documentclass[letterpaper, 10 pt, conference]{ieeeconf}  
\IEEEoverridecommandlockouts                              

\usepackage[pass,letterpaper]{geometry}

\makeatletter
\let\NAT@parse\undefined
\makeatother
\usepackage[square, sort&compress, comma, numbers]{natbib}

\usepackage{graphicx}        
\usepackage{multicol}        
\usepackage[bottom]{footmisc}
\usepackage{tikz}
\usepackage{subfigure}
\usepackage{caption} 
\usepackage{amsmath}
\usepackage{amssymb}
\usepackage{psfrag}
\usepackage{multirow}
\usepackage{url}
\usepackage{algorithmic}
\usepackage{algorithm}
\usepackage{array}
\usepackage{bm}
\usepackage{xspace}
\usepackage{color}
\usepackage{balance}

\DeclareGraphicsExtensions{.eps} 
\graphicspath{{Figures/}{./}}
\usepackage[bookmarks=true]{hyperref}
\usepackage[hyphenbreaks]{breakurl}

\title{A Fleet of Miniature Cars for Experiments in Cooperative Driving}

\author{Nicholas Hyldmar$^*$, Yijun He$^*$, Amanda Prorok
\thanks{All authors are with the University of Cambridge, UK: {\tt\small \{nh490, yh403, asp45\}@cam.ac.uk}. $^*$These authors contributed equally to this work. We gratefully acknowledge the Isaac Newton Trust who are supporting Amanda Prorok through an Early Career Grant.} %
}

\begin{document}

\maketitle
\thispagestyle{empty}
\pagestyle{empty}

\begin{abstract}
We introduce a unique experimental testbed that consists of a fleet of 16 miniature Ackermann-steering vehicles. We are motivated by a lack of available low-cost platforms to support research and education in multi-car navigation and trajectory planning. This article elaborates the design of our miniature robotic car, the \emph{Cambridge Minicar}, as well as the fleet's control architecture. Our experimental testbed allows us to implement state-of-the-art driver models as well as autonomous control strategies, and test their validity in a real, physical multi-lane setup. Through experiments on our miniature highway, we are able to tangibly demonstrate the benefits of cooperative driving on multi-lane road topographies. Our setup paves the way for indoor large-fleet experimental research.
\end{abstract}


\input{sec_introduction.tex}

\input{sec_design.tex}

\input{sec_architecture.tex}
\input{sec_model.tex}

\input{sec_algorithms.tex}

\input{sec_experiments.tex}

\input{sec_conclusion.tex}


\newpage
\small{
\bibliographystyle{abbrv}
\bibliography{Bibliography}
}

\end{document}

%% file: sec_introduction.tex
\section{Introduction}
The deployment of connected, automated, and autonomous vehicles presents us with transformational opportunities for road transport. To date, the number of companies working on this technology is substantive, and growing~\cite{cbsreport}. 
Opportunities reach beyond single-vehicle automation: by enabling groups of vehicles to jointly agree on maneuvers and navigation strategies, real-time coordination promises to improve overall traffic throughput, road capacity, and passenger safety~\cite{dressler:2014,ferreira2010self}. However, coordinated driving for intelligent vehicles still remains a challenging research problem, due to unpredictable vehicle behaviors (e.g., non-cooperative cars, unreliable communication), hard workspace limitations (e.g., lane topographies), and constrained kinodynamic capabilities (e.g., steering kinematics, driver comfort).

Developing true-scale facilities for safe, controlled vehicle testbeds is massively expensive and requires a vast amount of space. For example, the University of Michigan's MCity Test Facility cost US\,\$10 million to develop and covers 32 acres (0.13 km$^2$). As a consequence, using fleets of actual vehicles is possible only for very few research institutes worldwide. Moreover, although such facilities are excellently suited for research and development, safety and operational concerns prohibit the integration of educational curricula and outreach activities.
One approach to facilitating experimental research and education is to build low-cost testbeds that incorporate fleets of down-sized, car-like mobile platforms. Following this idea, we propose a multi-car testbed that allows for the operation of tens of vehicles within the space of a moderately large robotics laboratory, and allows for the teaching and research of coordinated driving strategies in dense traffic scenarios. 

Our motivation is to design a testbed that scales to a large number of cars so that we could test vehicle-to-vehicle interactions (cooperative as well as non-cooperative) and the effect of these interactions in multi-car traffic scenarios. 
Although a number of low-cost multi-robot testbeds exist (e.g., ~\cite{pickem:2017,jimenez:2013testbeds}), most use robotic platforms with differential drive kinematics (which tend to also have limited maximum speeds). Few testbeds integrate car-like Ackermann-steering vehicles, and in very small numbers (e.g.,~\cite{king:2004coordination}).

In this work, we propose the design of a low-cost miniature robotic car, the \emph{Cambridge Minicar}, which is based on a 1:24 model of an existing commercial car. The Minicar is built from off-the-shelf components (with the exception of one laser-cut piece), and costs approximately US\,\$76 in its basic configuration. Its low cost allows us to compose a large fleet, which we use to test navigation strategies and driver models.
Overall, our contributions in this work are \emph{(a)} the design of a low-cost miniature robotic car, \emph{(b)} the proposition of a system architecture that incorporates decentralized multi-car control algorithms for indoor testing of real large fleets, and \emph{(c)} the availability of our designs and code in an open-source repository~\footnote{\url{https://github.com/proroklab/minicar}}.

\begin{figure}
\centering
\includegraphics[width=0.98\columnwidth,natwidth=800,natheight=450]{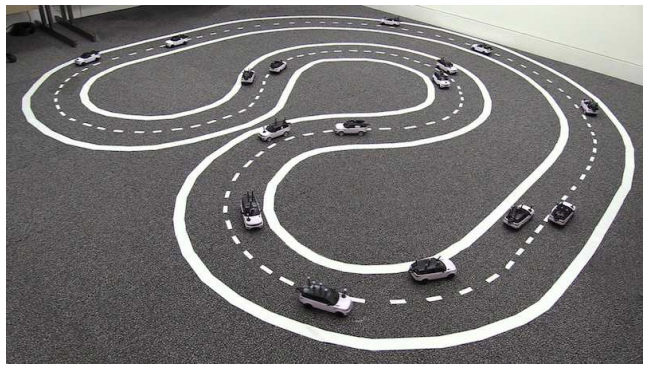}
\caption{The fleet of Minicars on a U-shaped two-lane miniature freeway. The inner and outer track lengths are 16\,m and 17\,m, respectively.
\label{fig:minicar_fleet}}
\end{figure}

\section{Existing Platforms}
A variety of low-cost mobile robot platforms are available for research and education. The work in~\cite{paull:2017} presents a very recent comprehensive overview of platforms that cost less than US \$\,300, have been designed in the last 10 years, or are currently available on the market.
Apart from the Kilobot~\cite{rubenstein:2012} and the AERobot~\cite{rubenstein:2015}, which have slip-stick forwards motion, all other robots in this overview are wheeled differential drive platforms~\cite{wilson:2016,mclurkin:2014,paull:2017,scribbler3,kernbach:2011,riedo:2013,dekan:2013,hemisson}. The lack of low-cost, Ackermann-steering robots is apparent.

There are a few recent robot designs that are based on Ackermann-steering platforms, and we provide an overview thereof in Table~\ref{tab:platforms}. 
The largest of these platforms is the Georgia Tech AutoRally~\cite{williams:2016}\footnote{\scriptsize{{\url{https://autorally.github.io/}}}}. It is based on a 1:5-scale vehicle chassis, runs on an ASUS Mini-ITX motherboard that includes a GPU, and is capable of fast autonomous driving using on-board sensors only. It is specifically designed for outdoor experimentation and the testing of aggressive maneuvers (all computational components are housed in a rugged aluminum enclosure able to withstand violent vehicle rollovers.) 
The MIT Racecar~\cite{karaman:2017}\footnote{\scriptsize{{\url{https://github.com/mit-racecar}}}} and the Berkeley Autonomous Race Car (BARC)~\cite{gonzales:2016}\footnote{\scriptsize{{\url{http://www.barc-project.com/}}}} are 1:10-scale rally cars based on a chassis that is commercially available (Traxxas). Similar to the AutoRally, these platforms feature high-end computational units with GPUs that allow for the addition of numerous sensors (e.g., laser range finder, camera, GPS) for on-board autonomy. As listed in Table~\ref{tab:platforms}, the price of these three platforms lies in the range \$\,800-\$\,9300 for the core alone (chassis, motors, and computational units). Due to the significant cost and moderately large size of the platforms, it is difficult to facilitate indoor experiments that include large numbers of these vehicles. 

At the other end of the spectrum lies the ETHZ ORCA Racer~\cite{liniger:2015}\footnote{\scriptsize{{\url{https://sites.google.com/site/orcaracer/home}}}}, which is based on a 1:43-scale platform. The chassis is provided by a Kyosho \emph{dnano} RC race car (achieving top speeds of more than 3 m/s). In order to control the platform, the authors designed a custom PCB with an ARM Cortex M4 microcontroller, a Bluetooth chip, and H-bridges for DC motor actuation (for the steering servo and drive train). The overall cost of one car is \$\,470 (approximately \$\,300 for the PCB parts and \$\,170 for the \emph{dnano} chassis).
This platform provides interesting dynamic capabilities, and a large number of cars can be operated in very confined spaces. Although the platform is considerably cheaper than the aforementioned models, it is still significantly more expensive than our Minicar (with similar capabilities). Furthermore, the platform is too small to carry an off-the-shelf computer, such as Raspberry Pi Zero W~\footnote{\scriptsize{\url{https://www.raspberrypi.org}}} (which facilitates communication via WiFi, as well as the addition of various sensors). Finally, its design and accompanying software tools are not open-sourced. These factors limit the extensibility and scalability of the testbed.

\begin{table}[tb]
\centering
\begin{tabular}{l|l|l}
\hline
\textbf{Car} & \textbf{Scale} & \textbf{Price (USD) }\\ \hline
ETHZ ORCA Racer~\cite{liniger:2015} & 1:43 & \$\,470 \\
\textbf{Cambridge Minicar} & 1:24 & {\$\,76.5} \\ 
BARC~\cite{gonzales:2016} & 1:10 & \$\,840 \\
MIT Racecar~\cite{karaman:2017} & 1:10 & \$\,1060 \\
GATech AutoRally~\cite{williams:2016} & 1:5 & \$\,9210 \\
\hline
\end{tabular}
\caption{Overview of Ackermann-steering platforms.\protect\footnotemark
\label{tab:platforms}}
\end{table}
\footnotetext{This overview lists platforms developed in the last 10 years. Prices indicate the cost of the core body, including chassis, motors and computational units (excluding additional sensor components.)}

%% file: sec_design.tex
\section{Vehicle Design}
\begin{figure}
\centering
\includegraphics[width=0.98\columnwidth]{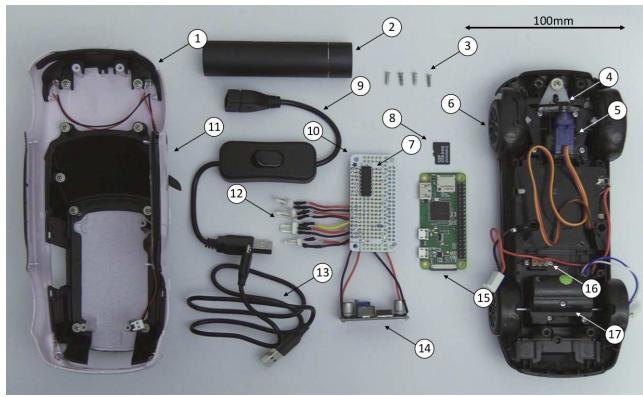}
\caption{Exploded view of a Cambridge Minicar. 1:\,Headlights 2:\,Portable charger 3:\,Casing screws 4:\,Gear 5:\,Servo 6:\,Lower casing 7:\,H-bridge 8:\,Micro SD card 9:\,USB switch 10:\,Circuit board 11:\,Upper casing 12:\,JST connectors 13:\,Micro USB cable 14:\,Boost converter 15:\,Raspberry Pi Zero W 16:\,Motor switch 17:\,Drive motor
\label{fig:explode_view}}
\end{figure}

The main components of the Minicar are a Raspberry Pi Zero W, a chassis with forwards drive train and servo-motors, and two battery sets. Figure~\ref{fig:explode_view} shows an exploded view of the Minicar. 
It is $75\times81\times197$\,mm and weighs 450\,g (including batteries). The logic can be powered for over 5 hours and the motors for 2.2 hours at 0.3\,m/s.
The logic is powered separately from the motors to isolate it from the noisy environment and increase the car's runtime. The motor and logic powers have capacities of 2500\,mAh and 3350\,mAh respectively. The AA batteries supply 4.2\,V to the servo which is increased to 7\,V for the motor by the boost converter. The servo logic and motor enable pin are controlled using pulse width modulation (PWM) by the Pi Zero W. A servo arm is connected to a laser-cut gear that meshes with the existing steering gear.
The vehicle's wheel base is $L=122$\,mm. Its minimum turning radius is $R=$ 0.56\,m, its maximum steering angle is $|\psi|_{\mathrm{max}} = 18^{\circ}$, its maximum steering rate is $|\dot{\psi}|_{\mathrm{max}}=$ $0.076$\,rad/s and its maximum forwards speed is $v_{\mathrm{max}} =$ 1.5\,m/s (this can be increased by tuning the output voltage of the boost converter).

Table~\ref{tab:minicar} lists the Minicar's components.
The cost can be reduced by \$\,28 by using power sources of lower capacity, removing the logic switch and JST connectors and replacing the board with a stripboard. The process of making a Minicar involves modifying the casing, soldering the circuit board, printing the gear, and fixing the components in place. This takes approximately 3 hours per car.


\begin{table}[tb]
\centering
\begin{tabular}{l|l|l}
\hline
\textbf{Component} & \textbf{Item} & \textbf{Price (USD) }\\
\hline
Computation & Raspberry Pi Zero W & \$\,12.3 \\
Memory & 8GB Micro SD card & \$\,5.2 \\
Chassis \& Motor & 1:24 Range Rover Sport & \$\,15.6 \\
Steering & Micro-servo & \$\,6.2  \\
Motor Power & AA Batteries & \$\,11.1 \\
Logic Power & Portable Charger & \$\,15.6 \\
Boost converter & XL6009 & \$\,2.6 \\
H-Bridge & L293D & \$\,0.5 \\
Board & Proto Bonnet & \$\,4.5 \\
Logic Switch & USB Switch & \$\,2.9 \\
\hline
\end{tabular}
\caption{Cambridge Minicar: overview of components.
\label{tab:minicar}}
\end{table}

%% file: sec_architecture.tex
\section{Testbed}

\begin{figure}[tb]
\centering
\includegraphics[width=0.9\columnwidth]{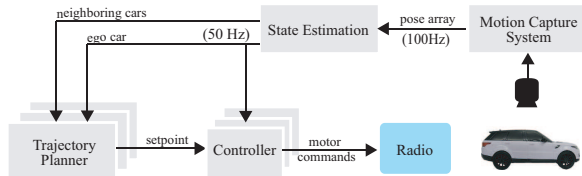}
\caption{Diagram of our system architecture.}
\label{fig:system_architecture}
\end{figure}

The architecture of our system is illustrated in Figure~\ref{fig:system_architecture}. We use an OptiTrack motion capture system based on passive reflective markers to provide real-time feedback on the vehicles' positions. Each Minicar is equipped with a unique configuration of five markers. The motion capture system tracks the Minicars and provides pose measurements at 100\,Hz. An extended Kalman filter uses the pose array to provide an estimate of each vehicle's state that includes vehicle pose $[x, y, \theta]$, velocity $v$ and steering angle $\psi$.
We implement an inner and outer loop control method for trajectory tracking. 
The outer loop (i.e., Trajectory Planner) generates trajectories (or uses pre-computed trajectories, such as freeway lanes) to compute velocity and steering angle setpoints, which are fed to the inner loop. In Section~\ref{sec:algorithms}, we provide an example of a Trajectory Planner for use in multi-lane freeway traffic. 
The inner loop (i.e., Controller) is responsible for correcting motor commands using PID control on velocity and steering angle. These control values are sent to the vehicles over broadband radio. The on-board computer applies the corresponding motor commands with pulse-with modulated signals. At the scale of our current system (16 vehicles), we did not notice any communication latency. Should this become an issue at larger numbers of vehicles, we note that the architecture can easily be scaled by considering multi-radio solutions, such as in~\cite{crazyswarm2017}.

Our testbed architecture is designed for ease of use, and our key aim is the rapid development and testing of driving behaviors on car-like robots (such as the Minicar).
Although the Minicar's design allows for the integration of sensors (such as an IMU or camera~\footnote{A dedicated Raspberry Pi Camera Module costs less than \$\,30; its miniature form factor allows it to be easily fitted onto the Minicar.}), and its on-board computer is capable of performing its own state estimation and control computations, we decided to keep the intelligence off-board, emulating proprioceptive and exteroceptive observations through the motion capture system instead.
In our setup, a workstation runs a thread for each Minicar's Trajectory Planner and Controller. Upon booting, the Minicar executes a listener that waits for motor commands. This software design choice facilitates rapid testing, and allows us to focus on the development of driving strategies for large Minicar fleets.
Our fleet operates on a miniature two-lane U-shaped freeway, shown in Figure~\ref{fig:minicar_fleet}.

%% file: sec_model.tex
\section{Trajectory Tracking}
\begin{figure}
\centering
\psfrag{R}[cc][][0.8]{$\mathbf{x}_d$}
\psfrag{Q}[cc][][0.8]{Q}
\psfrag{P}[cc][][0.8]{P}
\psfrag{y}[cc][][0.8]{$y$}
\psfrag{x}[cc][][0.8]{$x$}
\psfrag{u}[cc][][0.8]{$\theta$}
\psfrag{p}[cc][][0.9]{$\psi_d$}
\psfrag{q}[cc][][0.9]{$\psi$}
\psfrag{1}[cc][][0.8]{$l_1$}
\psfrag{2}[cc][][0.8]{$l_2$}
\psfrag{a}[cc][][0.8]{$\mathbf{x}_t$}
\psfrag{t}[lt][][0.8]{Reference Trajectory}
\psfrag{i}[lc][][0.8][4]{Reference Car}
\psfrag{e}[cc][][0.8][20]{Real Car}
\includegraphics[width=0.7\columnwidth]{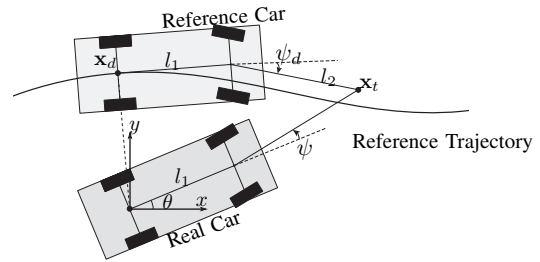}
\caption{Illustration of our trajectory tracking strategy, based on~\cite{linderoth2008nonlinear}.
\label{fig:tracking_model}}
\end{figure}

The Minicar has Ackermann steering geometry, and its kinematics can be approximated by the bicycle model, with motion equations as follows:
\vspace{-0.1cm}
\begin{eqnarray}\label{eq:equation_of_motion}
\dot{x} &=& v\cos{\theta} \\
\dot{y} &=& v\sin{\theta} \\
\dot{\theta} &=& L^{-1}{v\tan{\psi}}
\end{eqnarray}
where $\psi$ is the steering angle, $v$ is the forward speed, and $L$ is the vehicle's wheel base.
We use the lateral control strategy introduced in~\cite{linderoth2008nonlinear}. This strategy has the advantage of being speed independent, so that the velocity of the vehicle can be controlled independently for other purposes and still converge to the desired pose on the trajectory.

Figure~\ref{fig:tracking_model} illustrates the control strategy. The real vehicle with yaw $\theta$ and position $\mathbf{0}$ is projected orthogonally onto the reference trajectory to create a virtual reference vehicle at point $\mathbf{x}_d$, which is the closest point on the reference trajectory to the real vehicle. The yaw of the virtual reference vehicle is aligned with the tangent at $\mathbf{x}_d$, and its steering angle $\psi_d$ is such that the vehicle's turning radius coincides with the curvature $\kappa$ of the reference trajectory at $\mathbf{x}_d$.
We compute the target point $\mathbf{x}_t$ with
\begin{eqnarray}
\label{eq:equation_of_motion}
x_t &=& x_d + l_1 \cos \theta_d + l_2 \cos(\theta_d + \psi_d) \\
y_t &=& y_d + l_1 \sin \theta_d + l_2 \sin(\theta_d + \psi_d)
\end{eqnarray}
where the steering angle of the reference vehicle is $\psi_d = \arctan(l_1 \kappa)$.
The real vehicle's steering angle is
\begin{equation}\label{eq:steering}
\psi = \mathrm{arctan2}(y_t-l_1 \sin \theta,x_t-l_1 \cos \theta)-\theta.
\end{equation}
As noted in~\cite{linderoth2008nonlinear}, the choice of parameters $l_1$ and $l_2$ affect the control; small values for $l_1$ lead to fast control, but may lead to an overshoot of the reference trajectory; small values for $l_2$ (gain) lead to fast control, but also mean high control values, which may lead to a saturation of the steering angle. We experimentally tuned our parameters, and set $l_1 = L$ and $l_2 = 2.3 L$.

%% file: sec_algorithms.tex
\section{A Multi-Car Traffic System}
\label{sec:algorithms}

Building on the system architecture and vehicle control strategy described above, we design an experimental multi-car multi-lane traffic system that emulates freeway driving in an indoor lab setup. In our architecture, each vehicle is controlled by its individual, independent trajectory planning module, and hence, heterogeneous driving behaviors are possible. Our aim is to show how our experimental setup allows us to validate various driving controllers on actual platforms in a freeway-like setting.
To this end, we implement three control strategies: \emph{(i)} the car is driven by an egocentric, human-like policy, \emph{(ii)} the car is driven by a cooperative policy, and \emph{(iii)} the car is driven by a human player (i.e., a gamified policy).

\begin{figure}[tb]
\centering
\psfrag{T}[lc][][0.8]{\textbf{Trajectory Planner}}
\includegraphics[width=0.9\columnwidth]{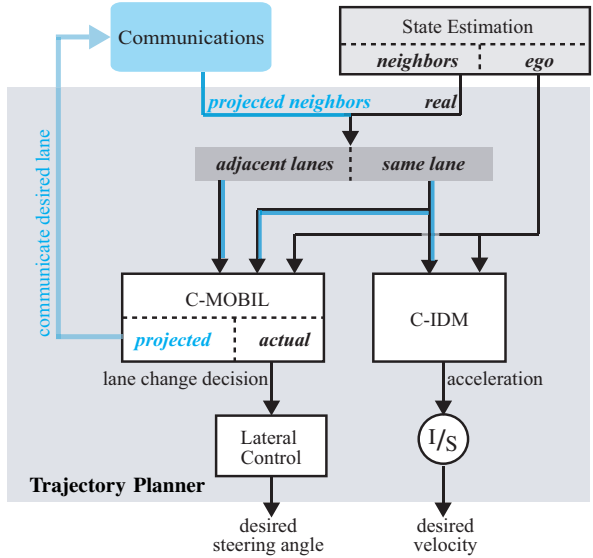}
\caption{Diagram of trajectory planner with our two main algorithmic modules, C-MOBIL and C-IDM. The ego vehicle plans a trajectory and velocity profile based on its own state, and on the actual state of its neighboring vehicles. In a cooperative approach (with diagram elements marked in blue), the algorithm modules also integrate the projected (desired) states of neighboring vehicles.}
\label{fig:planner}
\end{figure}

\subsection{Egocentric Driving}
\label{sec:ego_driving}
Our human-like driving model treats longitudinal and lateral control as two independent entities. It uses an acceleration model that controls longitudinal motion along the current car lane, and a steering model that controls lateral motion across multiple lanes.

\textbf{Longitudinal Control.}
Our longitudinal control is based on the Intelligent Driver Model (IDM) first proposed in~\cite{treiber2000congested}. The idea underpinning IDM is that a vehicle's acceleration is a function of the vehicle's current velocity $v$, it's gap $s$ to the preceding vehicle, and the approach rate $\Delta v$ to the preceding vehicle. It is formalized as
\begin{equation}
    a_{\mathrm{IDM}} = \alpha \left [  1- \left ( \frac{v}{v_0} \right)^\delta - \left (\frac{s^{\star}(v,\Delta v)}{s} \right )^2  \right ]
\end{equation}
where $s^{\star}$ is a function determining the desired minimum gap to the preceding vehicle. We compute this value as
\begin{equation}
    s^{\star}(v,\Delta v) = s_0 + T v + \frac{v \Delta v}{2 \sqrt{\alpha \beta}}.
\end{equation}
where $s_0$ corresponds to a jam distance.
The vehicle's control input at time $t$ is computed through the integrator $v_t = a_{\mathrm{IDM}} \cdot \Delta t + v_{t-1}$.

\begin{table}[tb]
\centering
\begin{tabular}{l|l|l}
\hline
\textbf{IDM Parameter} & \emph{Normal} & \emph{Aggressive}\\ \hline
Desired velocity $v_0$ [m/s] & 0.4 & 0.4 \\ 
Time headway $T$ [s] & 2.0 & 2.0 \\ 
Max. acceleration $\alpha$ [m/s$^2$] & 0.5 &  1.0 \\ 
Desired deceleration $\beta$ [m/s$^2$] & 0.3 & 0.5 \\ 
Acceleration exp. $\delta$ & 4  &  4 \\ 
Jam distance $s_0$ [m] & 0.1 &  0.1   \\ \hline 
\textbf{MOBIL Parameter} &  \emph{Normal} & \emph{Aggressive} \\ \hline
Politeness $p$ & 0.5 & 1.0 \\
Safe breaking $\beta_n$ [m/s$^2$] & $0.7 \cdot \alpha$ & $0.7 \cdot \alpha$ \\
Accel. thresh. $\Delta a_T$ [m/s$^2$]  & 0.4 & 0.2\\
\hline
\end{tabular}
\caption{Minicar parameter values for lateral and longitudinal control, in a \emph{normal} and an \emph{aggressive} mode.
\label{tab:IDM}}
\end{table}

\textbf{Lateral Control.}
Our lateral control strategy builds on the MOBIL lane changing policy proposed in~\cite{kesting2007general}. The MOBIL strategy decides whether a vehicle should change a lane or not based on two steps. First, it ensures that a safety criterion is met, i.e., it guarantees that after the lane-change, the deceleration of the new follower vehicle $a_n$ does not exceed a safety limit, $a_n \geq \beta_n$. Second, it validates via an incentive criterion that checks whether the local traffic situation of the car would improve, given a lane change. This incentive model also involves the immediately affected vehicles, i.e., the new and old follower vehicles. We evaluate the difference in acceleration for the current vehicle ($\Delta a_c$), the new follower ($\Delta a_n$), and the old follower ($\Delta a_o$), whereby a positive acceleration change for the current car is considered favorable. A politeness factor $p \in [0,1]$ determines how much heed is payed to the new and old follower vehicles. The incentive to change lanes is controlled by a switching threshold $\Delta a_T$, which ensures that a certain advantage is achieved through the prospected lane-change maneuver. This is formalized by the following inequality:
\begin{equation}
    \Delta a_c + p (\Delta a_n + \Delta a_o) > \Delta a_{T}.
\end{equation}

The parameter values that we used for both IDM and MOBIL controllers are detailed in Table~\ref{tab:IDM}.

\textbf{Remarks.}
Testing the aforementioned models on our experimental testbed demonstrated the need for adaptations that take into account the physical vehicles' kinematic and dynamic constraints.

Firstly, we noticed the the jam distance $s_0$ requires an additional \emph{escape} distance, which is a function of the ego vehicle's desired speed and the front vehicle's current speed, $v_f$, in order facilitate lane merging. Hence, we used an effective jam distance $\hat{s}_0 = s_0 + s_e$, where $s_e$ follows the rule:
\[ 
s_e(v_f,v_0) = \left \{
  \begin{tabular}{l}
  0{,~~if $v_f/v_0 >1$}\\
  $2L \left [ 2 \left ( \frac{v_{f}}{v_0} \right )^3-3 \left ( \frac{v_{f}}{v_0} \right )^2+1 \right ]${,~~else.}
  \end{tabular}
\right.
\]
This ensures a larger escape distance when the speed of the front vehicle is small relative to the ego car's desired speed, and that this distance approaches zero as the speeds become equal.
Secondly, a lane change on real cars is not immediate, and takes time to be completed. Hence, during this transition, cars on the original lane will still consider the state of ego vehicle in their longitudinal control, until the ego vehicle has completely entered the new lane. Similarly, the ego car starts using information about the state of new preceding and following cars on the new lane, as soon as it starts changing lanes.
Finally, an extra safety constraint is added to the MOBIL model to check that the escape distance between the ego vehicle and the front vehicle is large enough to prevent crashing when changing lanes.

\subsection{Cooperative Driving}
\label{sec:coop_driving}
Our cooperative driving strategy builds on the assumption that vehicles within visibility range $c$ communicate with one another to share \emph{intended} maneuvers before actually executing them. This allows the vehicles to cooperate about lane-changing decisions, and hence, plan efficient paths that maximize traffic throughput, whilst ensuring safety.
To test our setup's capability of validating the effects of cooperative driving we implement an approach that builds on the following key ideas:
\begin{itemize}
    \item When the ego vehicle decides to change its lane, it communicates with its neighboring vehicles, projecting a virtual counterpart vehicle at the desired new state.
    \item Vehicles within communication range of the ego vehicle receive information about its projected (virtual) state. They take this information into account to accelerate or decelerate, as a function of their relative positions to the virtual vehicle. 
\end{itemize}
We implement this behavior by modifying the original IDM and MOBIL models as follows.

\textbf{\emph{Cooperative}-IDM (C-IDM).} Our cooperative-IDM model takes \emph{projected} vehicles into account in the form of weighted virtual vehicles. When a vehicle projects its desired state, the resulting virtual vehicle is given a weight $w_v$ between 0 and 1 depending on how urgent the lane-change is. This urgency is defined by $w_v = \min(1,\kappa (c-s))$, where $s$ is the gap between the actual ego vehicle (to whom the virtual vehicle belongs), and its preceding vehicle.
Based on this idea, our cooperative IDM model differs from the original IDM model in two ways. First, our model \emph{decelerates} when detecting a projected vehicle in front of it. It does this by returning an acceleration $a = \min(w_v \cdot \tilde{a}_{\mathrm{IDM}}, a_{\mathrm{IDM}})$, where $a_{\mathrm{IDM}}$ is the acceleration computed using an actual front vehicle (if present), and where $\tilde{a}_{\mathrm{IDM}}$ is the acceleration computed using the projected front vehicle. Second, it \emph{increases} the desired speed when in front of or too close to a virtual vehicle, so that it can make space for the desired, projected vehicle state of the ego vehicle that owns the virtual vehicle. This is formalized by the following rule:
\[
\tilde{v}_0=v_0 \left (1 + w_v \frac{c-\tilde{s}_{\mathrm{trail}}}{c} \right )
\]
where $\tilde{s}_{\mathrm{trail}}$ is the distance between the neighboring vehicle and the virtual vehicle behind it.

\textbf{\emph{Cooperative}-MOBIL (C-MOBIL).} 
Our cooperative-MOBIL model uses a higher safe braking acceleration value than the original model in order to pack vehicles more closely when changing lanes; we set $\beta_n = \alpha$. 
Also, we add a safety constraint to prevent crashing when the speeds are low and the acceleration given by IDM is low:
$s > s_0 + \gamma \Delta v$,
where $s$ is the gap between the ego vehicle and the front/rear vehicle on the nearby lane. The constant $\gamma$ corresponds to the time needed for an average lane change.


\subsection{Gamification}
We gamify our experimental setup by interfacing one car (or several cars) with a joystick or keyboard. This allows a human player to experience traffic amid different types of surrounding vehicle behaviors. It also allows us to stress-test the reactions of other cars in traffic to arbitrary maneuvers.

A user can choose between three modes of control: `manual', `semi-automatic' and `automatic'. `Manual' gives a user direct control over the speed and turning angle of the car. `Semi-automatic' confines the vehicle to the designated lanes and safety restrictions while allowing the user to select a maintained speed and change lanes. `Automatic' removes all control from the player and integrates the vehicle into the traffic.


%% file: sec_experiments.tex
\section{Experiments}
\label{sec:experiments}

We perform three sets of experiments on our vehicle fleet. Our aim is to \emph{(a)} demonstrate the navigation  capabilities of our Minicar, \emph{(b)} demonstrate the operation of the Minicar fleet, and \emph{(c)} show how the fleet is used to test and validate (novel) driving algorithms in a realistic, albeit miniature, setup.

\subsection{Trajectory Tracking}
\begin{figure}[tb]
\centering
\psfrag{x}[cc][][0.8]{X [m]}
\psfrag{y}[cc][][0.8][90]{Y [m]}
\psfrag{s}[lb][][0.7]{start}
\psfrag{a}[lc][][0.55]{Reference}
\psfrag{e}[lc][][0.55]{Minicar}
\includegraphics[width=0.65\columnwidth]{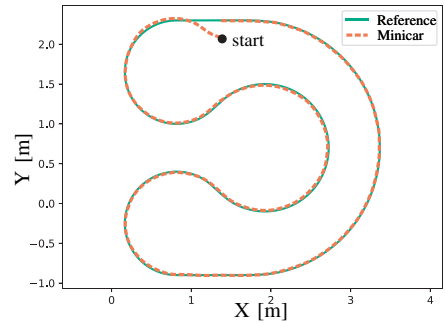}
\caption{Overhead representation of trajectory tracking accuracy for one loop of our U-shaped reference trajectory.\label{fig:tracking_accuracy}}
\end{figure}

We validate the Minicar's trajectory tracking capabilities. We run the Minicar on our U-shaped reference trajectory at a nominal speed of 0.4m/s.  Figure~\ref{fig:tracking_accuracy} shows an overhead plot of the tracking accuracy, for one loop on our U-shaped track. The average tracking error, measured over 7 loops of our track, is 14\,mm with a standard deviation of 6.3\,mm.

\begin{figure*}[h!]
\centering
\psfrag{o}[cc][][0.65][90]{Position along Track [m]}
\psfrag{T}[cc][][0.65]{Time [s]}
\psfrag{a}[cc][][0.65]{\textbf{\emph{Egocentric (Aggressive)}}}
\psfrag{n}[cc][][0.65]{\textbf{\emph{Egocentric (Normal)}}}
\psfrag{e}[cc][][0.65]{\textbf{\emph{Cooperative (Aggressive)}}}
\psfrag{i}[cc][][0.65]{\textbf{\emph{Cooperative (Normal)}}}
\psfrag{v}[cc][][0.65][90]{Velocity [m/s]}
\subfigure{\includegraphics[width=0.9\textwidth]{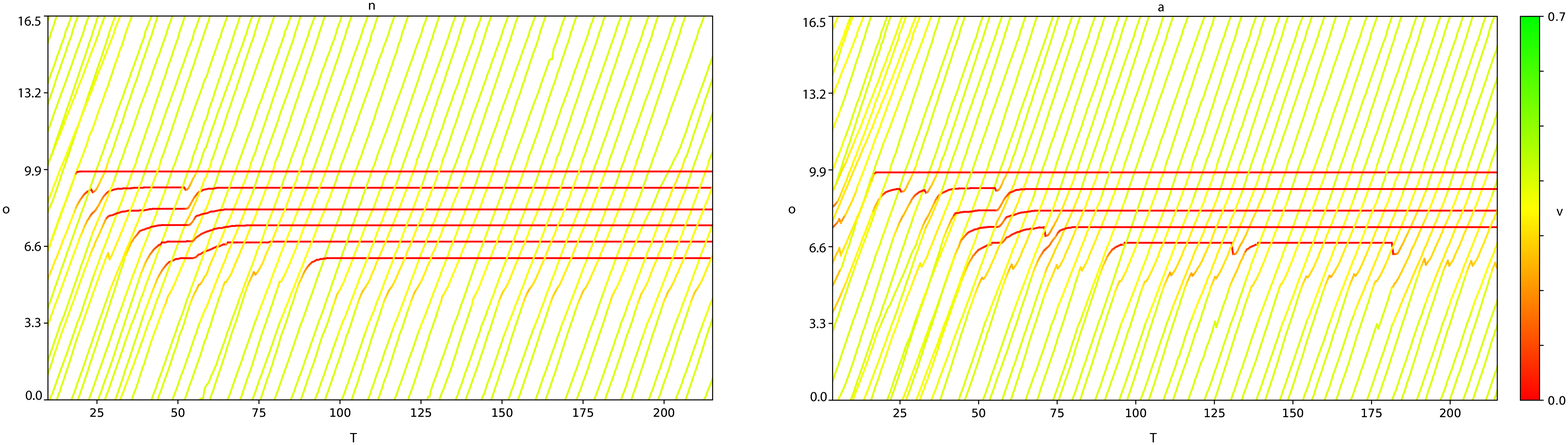}}
\subfigure{\includegraphics[width=0.9\textwidth]{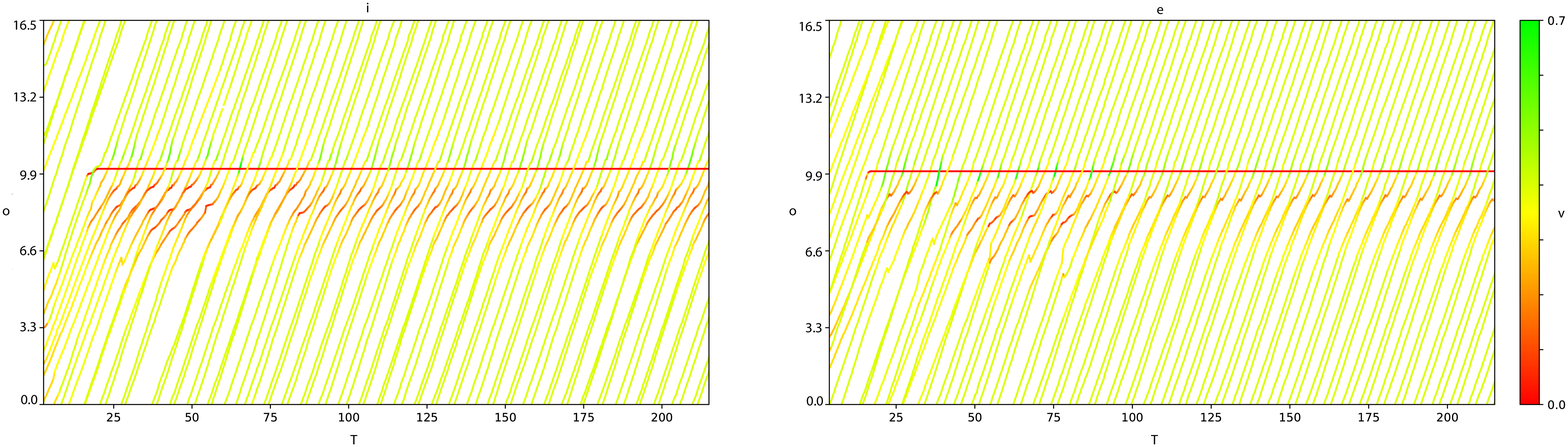}}
\caption{Results for experiments performed on 16 Minicars driving on our multi-lane U-shaped track, for \emph{egocentric} (top) and \emph{cooperative} (bottom) driving policies. The panels show vehicle positions the track, plotted as a function of time. One car is stopped after 20s to cause a traffic disturbance. The colorbar shows the velocity recorded. The panels on the left show data for the \emph{normal} parameter settings, panels on the right show the \emph{aggressive} parameter settings (see Table~\ref{tab:IDM}).
\label{fig:traffic}}
\end{figure*}

\subsection{Driving Behaviors in Multi-Car Traffic}
We implement four schemes to show the effect of different driving behaviors in multi-car traffic. We consider two driving policies: all cars are driven by \emph{(1)} the egocentric policy (described in Sec.~\ref{sec:ego_driving}), or \emph{(2)}, the cooperative policy (described in Sec.~\ref{sec:coop_driving}) with $c=2$\,m. For each policy, we use either the \emph{normal} or the \emph{aggressive} parameter set (see Table~\ref{tab:IDM}).
An experiment involves 16 Minicars driving on our U-shaped track and runs for 200s. At the start, the Minicars are evenly spaced out over the track. After 20s, one of the cars is told to stop (and hence, it blocks traffic on its current lane). The aim is to observe how the four different schemes react to this disturbance.

Figure~\ref{fig:traffic} shows the traffic flow for these four experiments, on the top row for the egocentric driving policy, and on the bottom row for the cooperative driving policy. In the egocentric scheme, we observe how the blockage creates a vehicle queue, which increases to 5 waiting vehicles at the latest stage. In contrast, the cooperative scheme overcomes vehicle queuing altogether (the traffic patterns exhibit very short stationary phases). With cooperative behavior, instead of queuing, a Minicar communicates its intention to lane-change; following vehicles in the new lane reduce their speeds to make space for this projected maneuver, hence maintaining traffic flow whilst ensuring safety. The overall throughput is improved in the cooperative scheme, and is further enhanced by aggressive control parameters; the average throughput values are reported in Table~\ref{tab:throughput}.


%

\begin{figure}[tb]
\centering
\psfrag{o}[cc][][0.65][90]{Position along Track [m]}
\psfrag{v}[cc][][0.65][90]{Velocity [m/s]}
\psfrag{T}[tc][][0.65]{Time [s]}
\psfrag{a}[rc][][0.5]{Accel.}
\psfrag{e}[rc][][0.5]{Turn}
\psfrag{n}[rc][][0.5]{Brake}
\psfrag{1}[lc][][0.5]{brake}
\psfrag{2}[lc][][0.5]{accel.}
\psfrag{3}[lc][][0.5]{left}
\psfrag{4}[lc][][0.5]{right}
\psfrag{5}[lc][][0.5]{stop}
\includegraphics[width=0.98\columnwidth]{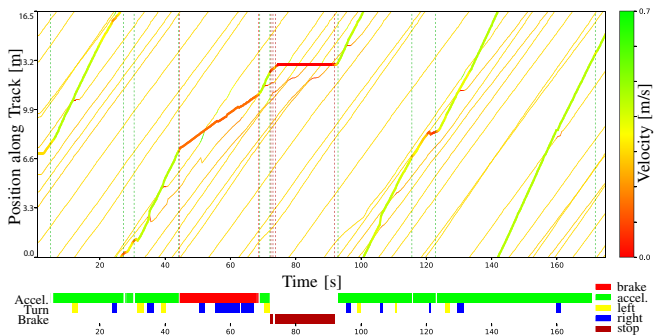}
\caption{A gamified experiment where a human controls one Minicar via a joystick, and 10 other Minicars are automated. The thick trajectory represents the played car. The colorbars along the time axis show the periods during which control commands are executed.
\label{fig:player}}
\end{figure}


\begin{table}[tb]
\centering
\begin{tabular}{l|l|l|l}
\hline
 & \textbf{Egocentric} &\textbf{Cooperative} & Improvement\\ \hline
\emph{Normal} & $0.245 \pm 0.036$ & $0.330 \pm 0.038$  & 35\% \\ 
\emph{Aggressive} & $0.277 \pm 0.045$ & $0.393 \pm 0.211$ & 42\% \\ 
\hline
\end{tabular}
\caption{Average throughput and standard deviation, measured in cars-per-second, for the four experiments shown in Figure~\ref{fig:traffic}. The third column shows the percent improvement of the cooperative scheme over the egocentric scheme.
\label{tab:throughput}}
\end{table}

\subsection{Gamification}
Figure~\ref{fig:player} shows an example of a player-controlled car amid 10 automated cars, in cooperative mode. We observe how the traffic is affected by the actions of the human player.

%% file: sec_conclusion.tex
\balance

\section{Discussion}
In this work, we provided the design of a fleet of miniature cars for research, education and outreach in the domain of automated multi-car systems. Our Ackermann-steering platform is one out of very few openly available designs; in particular, it fills a price-range gap, and is especially attractive for robotics labs that already possess telemetry infrastructure (such as motion capture). We propose a system architecture that is capable of integrating heterogeneous driving strategies (e.g., egocentric or cooperative) in a multi-lane setup. We demonstrate its applicability for large-fleet experimentation by implementing four different driving schemes that lead to quantifiable, distinct traffic behaviors. 

Although our current setup considers off-board intelligence, the platform is easily extended with an IMU and camera to provide full on-board autonomy. In future work, we plan to use our fleet for testing multi-car systems in more complex scenarios that include: \emph{(a)} road topographies with a larger number of lanes as well as intersections, \emph{(b)} heterogeneous vehicle behaviors in mixed traffic, and \emph{(c)} noisy sensing and delayed communications. Further research will investigate multi-objective optimization problems that also include driver comfort (as measured by vehicle accelerations).

